# Joint Forward-Backward Visual Odometry for Stereo Cameras

Raghav Sardana[1], Rahul Kottath[1,2], Vinod Karar[1,2], Shashi Poddar[1]
[1]CSIR-Central Scientific Instruments Organisation, Chandigarh - 160030, India,
[2]Academy of Scientific & Innovative Research, Ghaziabad - 201002, India
raghav.sardana@hotmail.com

## ABSTRACT

Visual odometry is a widely used technique in the field of robotics and automation to keep a track on the location of a robot using visual cues alone. In this paper, we propose a joint forward-backward visual odometry framework by combining both, the forward motion and backward motion estimated from stereo cameras. The basic framework of LIBVISO2 is used here for pose estimation as it can run in real-time on standard CPUs. The complementary nature of errors in the forward and backward mode of visual odometry helps in providing a refined motion estimation upon combining these individual estimates. In addition, two reliability measures, that is, forward-backward relative pose error and forward-backward absolute pose error have been proposed for evaluating visual odometry frameworks on its own without the requirement of any ground truth data. The proposed scheme is evaluated on the KITTI visual odometry dataset. The experimental results demonstrate improved accuracy of the proposed scheme over the traditional odometry pipeline without much increase in the system overload.

## 1 INTRODUCTION

Visual Odometry is one of the fundamental blocks in robotics research that helps in estimating camera motion using image frames. The image-based vehicle localization is necessitated by the unavailability of GPS signals in indoor, extra-terrestrial and under-surface environments for aiding inertial navigation [1]. The image-based motion estimation has been studied for several decades; however, the complexity and accuracy of these algorithms have improved exponentially with increasing processor speed. The feature-based VO pipeline has been detailed in Nister's [2] work, which uses preemptive based RANSAC approach for outlier rejection. In order to improve the performance of this scheme, several research works have been carried out in its pipeline such as design of better feature tracker, improved outlier rejection schemes, incorporating geometrical constraints, guided feature selection, etc.

Visual odometry can be classified differently such as monocular/stereo camera-based, feature/area-based, geometric/learning based [3]. Monocular VO methods using single camera suffers from the problem of scale estimation as compared to the stereo counterpart, which uses two cameras. The feature-based methods estimate motion by tracking interest points over image frames whereas the area-based schemes minimize a loss function defined over the complete image. Traditional methods that use camera geometry for motion estimation is explored widely even after the introduction of deep learning architectures. With the evolution of artificial intelligence, non-geometric learning based methods are also used to estimation motion while overcoming the problem of scale estimation in monocular VO [4]. However, the accuracy of non-geometric approaches is one of the main bottlenecks for using it in real time applications, and several recent works have been presented in the literature to address this issue [5, 6].

In this work, the motion information obtained by feeding the image pair in reverse order, that is, treating the next frame as the previous frame, and vice versa is incorporated in feature based VO pipeline. Although specific datasets are not available to explore the backward motion of the cameras, one can always feed the camera image inversely to the estimation framework for analysis. Lovegrove et.al [7] used the rear parking camera which looks backward at the ground to estimate vehicle motion by the complete image alignment approach. Pereira et al. [8] proposed a motion estimation scheme that processes the images in reverse order, taking advantage of sparse features moving away from the camera. They have shown that while using backward movement the features that are detected close to the camera have improved depth estimates, leading to better pose estimation. Recently, Yang et. al. [9] compared the performance of DSO [10] and ORB-SLAM [11] running in forward and backward mode. It was shown that ORB-SLAM performed better while running backward on some of the TUM Mono VO Dataset and termed this difference in forward and backward VO performance as motion bias. From the literature, it is found that very few works have explored this concept of feeding image frames inversely for motion estimation and this article is an attempt towards taking advantage of both the forward and backward motion sequences.

In this work, the forward and backward motion estimates between consecutive frames are combined in a single framework. It is shown that the inaccuracies occurring due to depth imprecision of new faraway points in incoming frames is complemented with the depth precision in incoming nearby points for the backward motion, improving the overall trajectory accuracy. The following sections provide further details on the implementation of joint forward-backward visual odometry pipeline which is simple and intuitive in nature as it directly averages the motion estimated from both the forward and backward motion. Section 2 provides the theoretical background of the various visual odometry pipeline stages, section 3 presents the proposed joint forward-backward visual odometry (JFBVO) approach, section 4 provides experimental analysis on KITTI dataset and, finally, section 5 concludes the paper.

## 2 THEORETICAL BACKGROUND

The feature-based visual odometry pipeline initiates with detecting and tracking feature points over consequent image frames. These feature points are triangulated to obtain their 3D locations, which are used to obtain the rotation and translation about all the three axes. The following two subsections provide a brief overview of feature detection, their matching, outlier rejection, and motion estimation subroutines.

### 2.1 Feature Detection and Matching

Features are those points in an image which are of specific interest and helps in providing sparse information of the image accurately. Although corner detectors such as Harris, Moravec, etc are very commonly used in the computer vision applications, they are not invariant to scale transformation. The scale-invariant feature transform (SIFT) technique proposed by Lowe [12] is a popular feature detector that formed the basis for further investigations on scale invariant feature detectors and is used in motion estimation schemes. Though these scale invariant feature detectors are robust, they are computationally complex and not always suitable for real-time systems. Therefore, customized blob and corner detectors are still used for several applications. Geiger et al. [13] employed a blob detector followed by non-maximum non-minimum suppression on the filtered images for reduced computational complexity. Once the features are detected, the corresponding features in the other image need to be matched with the help of distance measures such as the sum of squared difference (SSD), the sum of absolute difference (SAD), etc. In case of stereo images, circular matching among all four images is carried out (consecutive stereo pairs) to yield the set of features that is existent in all.

### 2.2 Motion Estimation & Outlier Rejection

Vision-based motion estimation is the process by which the rotation and translation of camera between two-time instants is assessed with the help of computer vision approach. Among different feature-based ego-motion estimation techniques, the 3D-to-2D technique as proposed in the LIBVISO2 framework is used here [13]. Considering, $x_i^l$ and $x_i^r$ be the 2D points in the current left and right image, respectively, and their corresponding 3D point in the previous left and right image as $X_i$, the reprojection error for 3D-to-2D method is given as

$$\sum_{i=1}^{N} \left\| x_i^l - \pi^l(X_i; r, t) \right\|^2 + \left\| x_i^r - \pi^r(X_i; r, t) \right\|^2 \quad (1)$$

Here, $\pi$ is the projection function that maps the 3D point to the 2D image point, and $r$, $t$ is the hypothesized rotation and translation that the 3D point follows. This estimated motion between consecutive frames is concatenated together to find the global pose with respect to the starting point. Once this motion is estimated with a few sub-sample of image points, the motion is verified over the complete set and checked for inlier percentage. This process of random sampling and consensus (RANSAC) check is repeated until a set with maximum inlier is obtained. It helps in removing outliers which can occur due to several reasons such as motion blur, occlusions, and illumination variation and improves estimation accuracy.

## 3 PROPOSED METHODOLOGY

In this paper, the conventional visual odometry pipeline has been improved by incorporating information from backward motion. In addition, two reliability measures have been proposed for evaluating motion estimated through visual odometry pipeline wherein ground truth trajectory is not available. Geiger et al. [13] proposed an ego-motion estimation framework LIBVISO2, with significant low computational complexity as compared to the other contemporary VO approaches. The features detected over image frames are matched circularly among the previous and current stereo image frames followed by an epipolar constraint with an error tolerance of 1 pixel. The reprojection error between the triangulated feature points and the corresponding 2D points on the image is minimized with the help of Gauss-Newton optimization and the outliers are removed using RANSAC to yield a convincing motion estimate. The LIBVISO2 framework is used here for further investigations and is termed as the conventional forward mode, wherein the previous frame and current frame are used as in the VO pipeline.

The feature points matched between two consecutive frames is fed inversely for the backward mode of VO scheme. In it, the previous frame is used as the current frame whereas the current frame is used as the previous frame, thus yielding another estimate of the same motion in the negative sense. In the proposed joint forward-backward motion estimation, the inverse of the backward motion is averaged with the motion estimated in forward mode to yield a robust estimate of motion. Let us consider $X_{k-1}$, $X_k$ as the sets of corresponding 3D points, $p_{l,k-1}$, $p_{r,k-1}$, $p_{l,k}$, $p_{r,k}$ as the location of 2D feature points in the left and right images at time instant *k-1* and *k,* respectively. The forward motion $T_f = [R_f | t_f]$ and backward motion $T_b = [R_b | t_b]$ can be computed by minimizing the reprojection error as shown in eq. (2) and (3).

$$\{R_f, t_f\} = arg\, min \sum_{i=1}^{N} \|p_{l,k}^i - \pi^l(R_f X_{k-1}^i + t_f)\|^2 \\ + \|p_{r,k}^i - \pi^r(R_f X_{k-1}^i + t_f)\|^2 \quad (2)$$

$$\{R_b, t_b\} = arg\, min \sum_{i=1}^{N} \|p_{l,k-1}^i - \pi^l(R_b X_k^i + t_b)\|^2 \\ + \|p_{r,k-1}^i - \pi^r(R_b X_k^i + t_b)\|^2 \quad (3)$$

Here $\pi^l$ and $\pi^r$ denote the projection functions of the left and right camera, superscript $i$ indicates $i^{th}$ feature point, and the argument to be minimized considers the error in location between 2D point in one image to the 2D point re-projected from the other image for both left and right camera and for all the corresponding points, $N$. The backward motion $T_b = [R_b|t_b]$ provides another estimate of the forward motion $T'_f = [R'_f|t'_f]$, i.e. $T'_f = T_b^{-1}$.

This helps in obtaining a refined estimate of the motion from $(k-1)^{th}$ frame to $k^{th}$ frame, $T_{k-1}^k = [R|t]$ by taking the Riemannian mean of $R_f$ and $R'_f$ [14] over the $SO(3)$ group of rotations and the mean of $t_f$ and $t'_f$ over $\mathbb{R}^3$ as

$$R = R_f(R_f^T R'_f)^{1/2} \quad (4)$$

$$t = \frac{t_f + t'_f}{2} \quad (5)$$

The overall flow diagram of the joint forward-backward estimation pipeline is shown in Fig. 1.

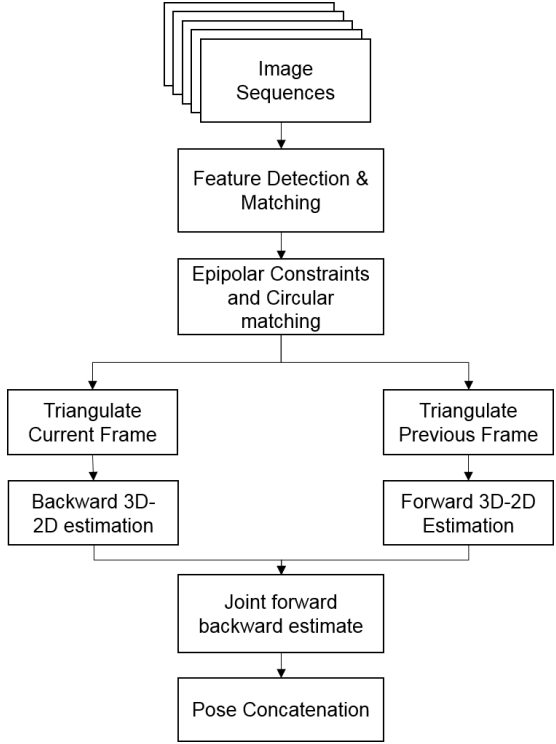

**Figure 1: Flow chart for joint forward-backward motion estimation**

The pose error metrics proposed here are relative and absolute in the sense that the relative error metric provides a frame to frame error only whereas the absolute error calculates the error at the current frame with respect to the origin. The forward-backward relative pose error metric measures the difference in pose estimated through forward and backward mode and is a self-reliability measure that does not require any ground truth trajectory. For any time instant $i$, let $F_i \in SE(3)$ and $B_i \in SE(3)$ be the absolute poses obtained in the forward and backward mode, respectively. The forward-backward relative pose error (FB-RPE) at time instant $i$ is defined as:

$$E_i^{rel} = (B_{i-1}^{-1} B_i)(F_{i-1}^{-1} F_i) = T_b T_f \quad (6)$$

For visualization, the magnitude of translation part of $\boldsymbol{E_i^{rel}}$ for the first sequence of the KITTI dataset is shown in Fig. 2. The figure depicts the small error value obtained by taking the norm of translation error for all the three axes with respect to time. For clarity of depiction, every $10^{th}$ image from the dataset is only considered for relative error value in the plot.

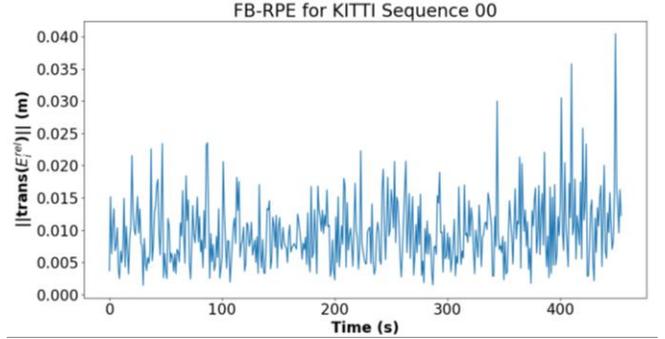

**Figure 2: Relative translation error between the forward and backward mode**

On a similar line to that of relative pose error, an absolute pose error is also defined here as a measure of the forward-backward consistency in the global sense. The forward-backward absolute pose error (FB-APE) at time instant $i$ is defined as:

$$E_i^{abs} = B_i F_i \quad (7)$$

The norm of the translation error for all three axes is plotted with respect to time in Fig. 3. For clarity of depiction, every $10^{th}$ image from the dataset is only considered for absolute error value in the plot.

As seen in Fig. 3, this error keeps on increasing as they keep growing over image frames. These relative and absolute metrics allow any visual odometry framework to self-evaluate its reliability at any given time instant.

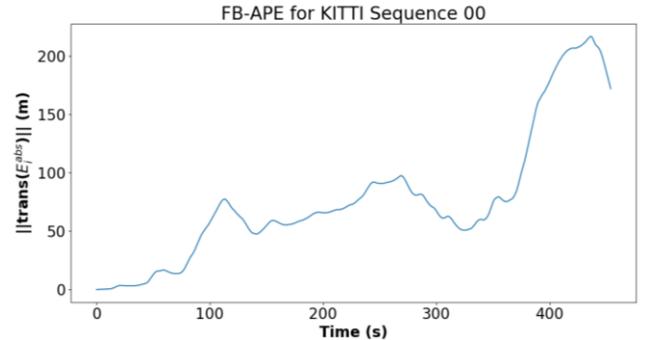

**Figure 3: Absolute translation error between the forward and backward mode**

## 4 RESULTS AND DISCUSSION

The proposed scheme of joint forward-backward visual odometry (JFBVO) is evaluated on the KITTI dataset to compare its performance against the traditional VO pipeline proposed by [13]. The KITTI odometry dataset [15] consists of 22 high-resolution stereo image sequences, divided into the training set, seq. 00 - 10 and test set, seq. 11-21. These sequences cover different scenarios like an urban city, highway and residential areas. The ground truth poses for the training dataset is available while the measurement accuracy for test sequences is obtained by submitting the obtained trajectory file on the KITTI website for evaluation. Table 1 provides the translation and rotational error for the LIBVISO2 pipeline run in forward mode, backward mode, and JFBVO mode. The average relative translation and rotation error achieved using the proposed JFBVO scheme for the training set is 1.35% and 0.36 (deg/100m), while for the test set, it is 1.43% and 0.38 (deg/100m), respectively. The mean of the rotation and translation error for both the training and test dataset depicts a significant improvement in estimated motion accuracy through the proposed JFBVO scheme as compared to the only forward and backward motion schemes.

**Table 1: Results on KITTI Dataset**

| Sequence | VISO2-Forward | | | VISO2-Backward | | | JFBVO | | |
|---|---|---|---|---|---|---|---|---|---|
| | $t_{rel}$ (%) | $r_{rel}$ (deg/100 m) | $t_{abs}$ (m) | $t_{rel}$ (%) | $r_{rel}$ (deg/100 m) | $t_{abs}$ (m) | $t_{rel}$ (%) | $r_{rel}$ (deg/100 m) | $t_{abs}$ (m) |
| 00 | 2.74 | 1.35 | 64.14 | 2.33 | 1.27 | 37.74 | 1.19 | 0.43 | 22.37 |
| 01 | 4.27 | 1.00 | 132.14 | 3.83 | 0.76 | 129.09 | 4.61 | 0.36 | 147.53 |
| 02 | 2.20 | 0.87 | 76.76 | 1.98 | 0.96 | 44.16 | 1.16 | 0.36 | 38.73 |
| 03 | 2.27 | 1.06 | 13.84 | 2.67 | 1.27 | 14.31 | 1.16 | 0.21 | 4.07 |
| 04 | 1.08 | 0.85 | 2.68 | 0.50 | 1.50 | 1.16 | 0.69 | 0.34 | 0.81 |
| 05 | 2.25 | 1.20 | 25.91 | 1.80 | 1.04 | 13.86 | 0.89 | 0.36 | 8.1 |
| 06 | 1.28 | 0.87 | 7.76 | 0.68 | 0.74 | 1.88 | 1.03 | 0.33 | 4.73 |
| 07 | 2.34 | 1.78 | 12.12 | 1.66 | 1.55 | 9.94 | 0.88 | 0.57 | 3.49 |
| 08 | 2.83 | 1.33 | 64.63 | 2.49 | 1.34 | 31.05 | 1.19 | 0.36 | 20.81 |
| 09 | 2.84 | 1.19 | 52.79 | 2.76 | 1.15 | 37.91 | 1.45 | 0.33 | 15.04 |
| 10 | 1.39 | 1.15 | 11.79 | 1.98 | 1.29 | 23.86 | 0.64 | 0.33 | 10.37 |
| **Mean (00-10)** | 2.32 | 1.15 | 42.23 | 2.06 | 1.17 | 31.36 | 1.35 | 0.36 | 25.09 |
| **Mean (11-21)** | 2.44 | 1.14 | - | 2.53 | 1.32 | - | 1.43 | 0.38 | - |

In order to evaluate the global consistency of the estimated trajectories, the absolute translation RMSE, $t_{abs}$ [16] values for all the three schemes has been provided in Table 1 for the training dataset. As the ground truth is provided in the reference frame of the left camera, no alignment procedures have been performed for computing the absolute translation error. The change in translation and rotation error averaged over all possible subsequences of length 100-800 meters is also a strong parameter to compare different methodologies in KITTI dataset. Accordingly, Fig. 4 depicts the overall error in the complete test set with respect to path length for all the three modes, that is, forward, backward and JFBVO. It can be observed that the translation error of both the forward and backward mode increases with respect to the path length whereas it remains nearly constant for JFBVO.

This evaluation has been done on an Intel i7-3770 CPU running at 3.40 GHz. The average single threaded runtime per image for JFBVO is 50 milliseconds while for LIBVISO2-forward/backward is 25 milliseconds. The simple forward/ backward mode takes lesser time for motion estimation at the cost of accuracy. For better visualization, the trajectories obtained for various sequences across different schemes are shown in Fig. 5. As seen, in Fig. 5 the trajectory for forward and backward mode seems complementary to each other and its advantage has been taken here through the proposed JFBVO scheme.

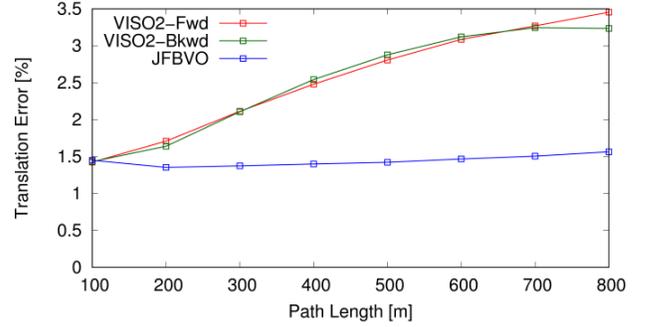

(a)

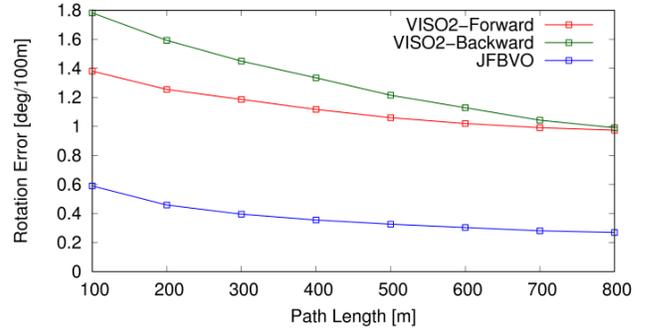

(b)

**Figure 4: (a) Relative translation error (percentage) and (b) rotation error (deg/100m) with respect to path length on the test set**

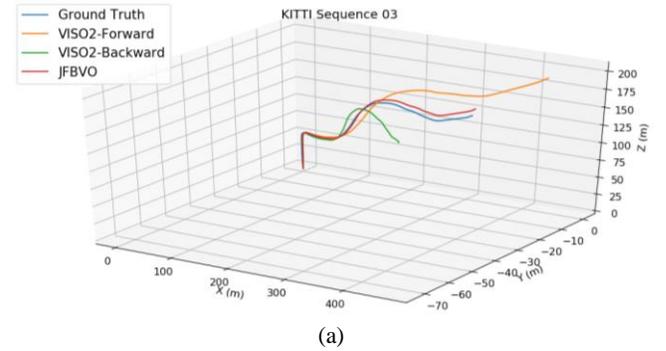

(a)

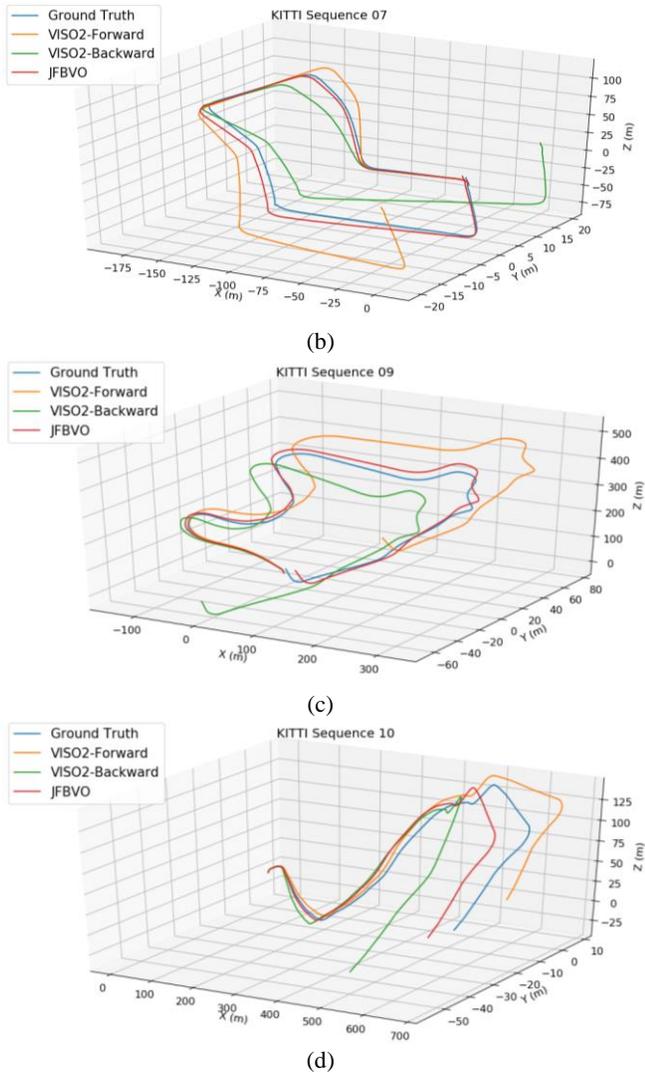

**Figure 5:** Trajectories obtained by VISO2-Forward, VISO2-Backward and JFBVO on KITTI (a) sequence 3, (b) sequence 7, (c) sequence 9 and (d) sequence 10

## 5　CONCLUSION

In this paper, a novel method of joint forward-backward visual odometry is proposed which incorporates cues from backward motion to improve the forward motion estimate. Forward and backward motion estimates between two image frames are computed at every time instant, which is then combined in an averaging framework. The LIBVISO2 algorithm is used here as the implementation baseline for evaluating the proposed algorithm. The proposed scheme of JFBVO is compared with the forward and backward mode of VO and evaluated on the KITTI dataset. The complementary effect of both the forward and backward VO provides a better estimate on combining both of them and can be inferred from the experimentation provided in this article. Added to this, two reliability measures, that is, forward-backward relative pose error and forward-backward absolute pose error metrics have been also proposed here to evaluate the reliability of motion estimate on its own while the ground truth data is not available. In the future, authors would also like to incorporate these error metrics in a learning framework to obtain reliable motion estimates.

## ACKNOWLEDGMENTS

This research has been supported by DRDO-Aeronautical Research & Development board through grant-in-aid project on design and development of visual odometry system.